\documentclass{article}

\usepackage{arxiv}
\usepackage{graphicx}
\usepackage[utf8]{inputenc} 
\usepackage[T1]{fontenc}    
\usepackage{hyperref}       
\usepackage{url}            
\usepackage{booktabs}       
\usepackage{amsfonts}       
\usepackage{amsmath}
\usepackage{amssymb}
\usepackage{nicefrac}       
\usepackage{microtype}      
\usepackage{lipsum}
\usepackage{accsupp}

\newcommand{\A}{\mathcal{A}}

\newcommand{\K}{\mathcal{K}}
\newcommand{\Emcon}{Repairs}
\newcommand{\Kb}{\mathcal{K}}
\newcommand{\Kbs}{\mathcal{K}s}

\newcommand{\KB}{\K = (\F, \R, \N)}
\newcommand{\F}{\mathcal{F}}
\newcommand{\R}{\mathcal{R}}
\newcommand{\N}{\mathcal{N}}

\newcommand{\Q}{\mathcal{Q}}

\newcommand{\Cl}{\mathcal{S}at}
\newcommand{\Clust}{Clust}
\newcommand{\Clo}{\mathcal{C}\ell}
\newcommand{\Dif}{\mathcal{DIF}}
\newcommand{\fresh}{Null}

\newtheorem{definition}{Definition}
\newtheorem{example}{Example}

\title{Distance-Based Approaches to Repair Semantics in Ontology-Based Data Access}

\author{
  César Prouté \\
  University of Montpellier\\
  \texttt{cesar.proute@etu.umontpellier.fr}\\
   \And 
  Bruno Yun\\
  University of Edinburgh\\
  \texttt{bruno.yun@ed.ac.uk}\\
   \And
 Madalina Croitoru \\
 University of Montpellier\\
  \texttt{croitoru@lirmm.fr} \\
}

\begin{document}
\maketitle

\begin{abstract}
In the presence of inconsistencies, repair techniques thrive to restore consistency by reasoning with several repairs. However, since the number of repairs can be large, standard inconsistent tolerant semantics usually yield few answers. 
In this paper, we use the notion of syntactic distance between repairs following the intuition that it can allow us to cluster some repairs ``close'' to each other. In this way, we propose a generic framework to answer queries in a more personalise fashion.
\end{abstract}


\section{Introduction}

In the knowledge representation and reasoning on the Semantic Web setting, the focus has been placed on a subset of first order logic called existential rules language because of its expressivity and decidability results \cite{gottlob_datalog+/-:_2014,cali_general_2009}. 
A knowledge base (KB) in this language is composed of both a layer of factual knowledge and a layer of ontological reasoning rules of deduction and negative constraints. 
The main reasoning feature in such a KB is answering queries over the knowledge deduced from the two aforementioned layers.
However, classical query answering approaches fail in the presence of contradictions.
Against this background, inconsistent-tolerant approaches have been developed to overcome this problem by restoring consistency \cite{benferhat_non-objection_2016,baget_inconsistency-tolerant_2016}.

Most of those query answering methods are based on maximally consistent subsets of the fact base (\emph{repairs}) that they manipulate using a modifier and an inference strategy \cite{baget_general_2016}.
However, for some applications, the use of all repairs may not be the most appropriate \cite{yun_inconsistency_2018}. 
In this paper, we propose to split the set of repairs into clusters so that the end-user can have a more personalised answer depending on the part of the database one is looking at, i.e.\ our approach allows the user to query on specific repairs of the database instead of considering the set of all repairs.

\section{Background notions}

The existential rules language \cite{cali_general_2009} is composed of formulae built with the usual quantifiers $(\exists,\forall)$, \textit{only} two connectors ($\rightarrow,\wedge$) and is composed of facts, rules and negative constraints.
A \textit{fact} is a ground atom of the form $p(t_1,\dots, t_k)$ where $p$ is a predicate of arity $k$ and $t_i,$ with $i\in [1, \dots, k]$,  constants. The set of all possible facts is denoted by $\A$. 
An existentially closed atom is of the form $\exists \overrightarrow{X} p(\overrightarrow{a}, \overrightarrow{X})$ where $p$ is a predicate, $\overrightarrow{a}$ is a finite set of constants and $\overrightarrow{X}$ is a finite set of existentially quantified variables.
An existential \textit{rule} is of the form $\forall \overrightarrow{X}, \overrightarrow{Y}$ $H[\overrightarrow{X},\overrightarrow{Y}] \rightarrow \exists \overrightarrow{Z} C[\overrightarrow{Z},\overrightarrow{X}]$ where $H$ (called the hypothesis) and $C$ (called the conclusion) are existentially closed atoms or conjunctions of existentially closed atoms and $\overrightarrow{X},\overrightarrow{Y},\overrightarrow{Z}$ their respective vectors of variables. A \textit{rule is applicable} on a set of facts $\F$ if and only if there exists a homomorphism \cite{baget_inconsistency-tolerant_2016} from $H$ to $\F$. Applying a rule to a set of facts (also called \textit{chase}) consists of adding the set of atoms of $C$ to the facts according to the application homomorphism.
Applying an existential rule will lead to the inference of new individuals represented with \textit{fresh variables} (also called nulls). For instance, applying the rule $\forall X person(X) \to \exists Y parent(Y,X)$ to $person(john)$ will create the atom $parent(\fresh_1)$ where $\fresh_1$ is a variable that has not been used before.
Different \textit{chase} mechanisms use different restrictions that prevent infinite redundancies \cite{baget_rules_2011}. Here, we use recognisable classes of existential rules where the chase is guaranteed to stop \cite{baget_rules_2011}.
A \textit{negative constraint} is a rule of the form $\forall \overrightarrow{X}, \overrightarrow{Y}$ $H[\overrightarrow{X},\overrightarrow{Y}] \rightarrow \bot$ where $H$ is an existentially closed atom or conjunctions of existentially closed atoms, $\overrightarrow{X},\overrightarrow{Y}$, their respective vectors of variables and $\bot$ is \emph{absurdum}. 


Let us consider two existentially closed conjunctions of atoms $F_1$ and $F_2$. We say that $F_1$ \emph{entails} $F_2$ denoted by $F_1 \models F_2$ if and only if there is a homomorphism from the set of atoms in $F_2$ to the set of atoms in $F_1$.
A conjunctive query is an existentially quantified conjunction of atoms. For readability, we restrict ourselves to Boolean conjunctive queries, which are closed formulas (the framework and the obtained results can be extended to general conjunctive queries). The set of all possible Boolean conjunctive queries is denoted by $\Q$.

\begin{definition} 
A \textit{KB} $\Kb$ is a tuple $\KB$ where $\F$ is a finite set of facts, $\R$ a set of existential rules and $\N$ a set of negative constraints. 
\end{definition}

\begin{example}
\label{ex:reunning-kb}
Let us consider the KB $\KB$ about two babies Milo ($m$) and Jane ($j$). We consider that there is a $siblings$ relation that is symmetric, a baby can either stay or go somewhere, but can only be in one place. If one sibling stays somewhere the other will too, if they both go to the same place then both will be happy, and finally if one goes somewhere one will get ill.

~\\

\begin{tabular}{cl}
$\F:$ & $\{ baby(m), go\_to(m, day\_care), go\_to(m, nanny), stay(m, home), baby(j),$\\
	&$ go\_to(j, day\_care), stay(j, home),siblings(m, j) \}$ \\
	&\\
$\R:$ & $\{ \forall X, Y (siblings(X, Y) \to siblings(Y, X)),$ \\
&$ \forall X,Y,Z (stay(X, Z), siblings(X, Y) \to stay(Y, Z)),$ \\
&$ \forall X,Z (go\_to(X, Z) \to get\_ill(X)),$ \\
&$ \forall X,Y,Z (go\_to(X, Z), go\_to(Y, Z), siblings(X, Y) \to happy(X), happy(Y)) \}$ \\
&\\
$\N:$ & $\{ \forall X (go\_to(X, nanny), go\_to(X, day\_care) \to \bot),$\\
&$ \forall X (go\_to(X, nanny), stay(X, home) \to \bot),$\\
&$ \forall X(go\_to(X, day\_care), stay(X, home) \to \bot) \}$
\end{tabular}
\end{example}

The set of all knowledge bases is denoted by $\Kbs$.
The \textit{saturation} of $\F$ by $\R$ is the process of entailing all atoms and conjunctions of atoms by using all rule applications from $\R$ over $\F$.
Please note that although the saturation may be infinite in the general case, we place ourselves in the Finite Expansion Set (FES) abstract class where the saturation is finite.
The output of this process is called the closure and is denoted by $\Cl_{\R}(\F)$.
A set $\F$ is said to be $\R$-\textit{consistent} if no negative constraint hypothesis can be entailed, i.e. $\Cl_{\R}(\F) \not \models \bot$. Otherwise, $\F$ is said to be $\R$-\textit{inconsistent}. The \emph{ground closure} of a set $\F$ by $\R$, denoted $\Clo_{\R}(\F)$, only keeps the ground atoms of the closure. Namely, $\Clo_{\R}(\F) = \{ $ground atoms $a \mid a \in \Cl_{\R}(\F)\}$. 
The notion of ground closure is widely used in the literature and especially when dealing with inconsistency tolerant inference since computing the intersection of closures is ambiguous\footnote{Atoms resulting from the application of rules with existential variables contain \emph{nulls} which makes the intersection of two closures undefined}.
As an example, the ICR approach is based on the intersection of the ground closure of all the repairs.
We assume that the existential rules are skolemized \cite{marnette_generalized_2009}, namely $\forall \overrightarrow{X}, \overrightarrow{Y}$ $H[\overrightarrow{X},\overrightarrow{Y}] \rightarrow \exists \overrightarrow{Z} C[\overrightarrow{Z},\overrightarrow{X}]$ will be replaced with $\forall \overrightarrow{X}, \overrightarrow{Y}$ $H[\overrightarrow{X},\overrightarrow{Y}] \rightarrow C[f(X),\overrightarrow{X}]$ where $f$ is a new symbol function.
Using the above mentioned methodology in our setting enables us to have the closure equal to the ground closure and guarantees a finite saturation.

A repair of a KB is a maximal $\R$-consistent for set inclusion subset of facts.

\begin{definition}[Repairs]
  Let $\KB$ be a KB, the set $R \subseteq \F$ is a repair if and only if $R$ is $\R$-consistent and there is no $R' \subseteq \F, R \subset R'$ such that $R'$ is $\R$-consistent.
  We denote the set of all repairs of $\Kb$ by $\Emcon(\Kb)$.
\end{definition}

\begin{example}[Cont'd Example \ref{ex:reunning-kb}]
The repairs of the KB $\Kb$ are:
\begin{itemize}
\item $r_0 = \{ baby(m), stay(m, home), baby(j), go\_to(j, day\_care) \}$
\item $r_1 = \{ baby(m), go\_to(m, day\_care), baby(j), stay(j, home) \}$
\item $r_2 = \{ baby(m), go\_to(m, nanny), baby(j), stay(j, home) \}$
\item $r_3 = \{ baby(m), go\_to(m, day\_care), baby(j), go\_to(j, day\_care), siblings(j, m) \}$
\item $r_4 = \{ baby(m), go\_to(m, nanny), baby(j), go\_to(j, day\_care), siblings(j, m) \}$
\item $r_5 = \{ baby(m), stay(m, home), baby(j), stay(j, home), siblings(j, m) \}$
\end{itemize}
\label{ex:list-repair}
\end{example}

In the OBDA setting rules and constraints act as an ontology used to ``access'' different data sources. These sources are prone to inconsistencies. As it is common in the literature, we suppose that the rules are compatible with the negative constraints, i.e. the union of those two sets is satisfiable \cite{lembo_inconsistency-tolerant_2010}. Indeed, the ontology is believed to be reliable as it is the result of a robust construction by domain experts. However, as data can be heterogeneous due to merging and fusion, the data is assumed to be the source of inconsistency. 

\section{Distance-based Inference framework $\Dif$}

In this section, we introduce the distance-based inference framework (DIF) and its three main components: the repair distance, the clustering function and the inconsistency tolerant inference.
It is similar to the works by \cite{konieczny_supported_2015} or  in \cite{yun_inconsistency_2018}, where only the ``best'' sets are used for reasoning.
While the the work in \cite{yun_inconsistency_2018} also selects the best repairs, they rely on the overall inconsistency of the facts of the knowledge base. In our approach, we propose to use a distance measure based on atoms similarity.
The section is organised as follows: in Section \ref{sec:dif-distance}, we introduce the notion of repair distance and recall Ramon's repair distance, in Section \ref{sec:dif-clustering}, we give examples of clustering functions and in Section \ref{sec:dif-inference}, we show how to modify inconsistency tolerant inferences.

Our framework is based on three layers. 
First, a \emph{repair distance} is used to calculate the distance between each pair of repairs in $\Emcon(\Kb)$. We will work with Ramon's repair distance, but any repair distance can be used.

\begin{definition}[Repair distance]
Let $\Kb$ be a KB, a repair distance (or a metric) on $\Kb$ is a function $d: \Emcon(\Kb) \times \Emcon(\Kb) \to \mathbb{R}^+$ if and only if it satisfies all the three following items:

\begin{enumerate}
\item $d(x,y) \geq 0 \text{ and } d(x, y) = 0 \iff x = y$, 
\item $d(x,y) = d(y,x)$ and,
\item $d(x,z) \leq d(x,y) + d(y,z)$.
\end{enumerate}

\end{definition}

Second, we need a \emph{clustering function}, i.e.\ a function that group repairs into sets, based on the values returned by a repair distance. This clustering function can be based on fixed distance threshold or other graph features in order to reach a desired number of clusters.

\begin{definition}[Clustering function]
Let $\Kb$ be a KB and $d_\Kb$ be the set of all repair distance on $\Kb$. A clustering function on $\Kb$ is a function $\text{\Clust}: \Emcon(\Kb) \times d_\Kb \rightarrow Part(\Emcon(\Kb))$ where $Part(\Emcon(\Kb))$ is the set of all partitions of $\Emcon(\Kb)$.
\end{definition} 

Third, we use an \emph{inconsistency tolerant inference relation} restricted to a cluster of repairs. At this step, one can use the usual inconsistency tolerant inference relations such as AR, IAR, ICR or the modifier-based semantics \cite{baget_general_2016}.

\begin{definition}[Inconsistency-tolerant inference]
An inconsistency-tolerant inference relation is a function $\models: \Kbs \times \Q \rightarrow \{True, False \}$. 
\end{definition}

Based on the previous notions, we define the DIF framework.

\begin{definition}[$\Dif$ framework]
Let $\Kb$ be a KB, a DIF on $\Kb$ is a tuple $\Dif = (d, \Clust, \models)$ where $d$ is repair distance on $\Kb$, $\Clust$ is a clustering function on $\Kb$ and $\models$ is an inconsistency tolerant inference.
%
\end{definition}

\subsection{$\Dif$ Distance}
\label{sec:dif-distance}

A repair distance is function that associates a positive value to each pair of repair.
This value is supposed to be lower the more the two considered repairs are ``similar to each other''.

In this section, we define the Ramon's repair distance by extending a distance function on first-order atoms \cite{ramon1998distance} to repairs with the \emph{matching distance} defined by \cite{ramon1998framework}.
The distance on first order atoms first assigns a size to an atom based on weights given to predicates and constants, and the frequency of apparition of the variables. The difference in size of two atoms defines a semi-distance $d_s$, it is not a proper distance because two distinct atom can have the same size. To remediate this, we use the \emph{least general generalisation} or \emph{lgg} between the two atoms. An atom $G$ is called the \emph{lgg} of $A$ and $B$ if and only if $G \models A$, $G \models B$ and for any atom $G'$ with $G' \models A$, $G' \models B$,
we have that $G' \models G$\footnote{If the least general generalisation of two atoms does not exist, we consider it as $\top$.}. Thus, the final distance is $d(A,B) = d_s(A,lgg(A,B)) + d_s(lgg(A,B), B)$.
In order to extend the distance $d$ between atoms to a distance between sets, we propose to take the least sum of the distances over all matchings between the atoms of the two sets.

\begin{example}[Cont'd Example \ref{ex:list-repair}]
Table \ref{tabl:dist-matrix-short} shows the distance matrix corresponding to the application of the Ramon's repair distance.

\begin{table}
\centering
\begin{tabular}{c|cccccc}
 & $r_0$ & $r_1$ & $r_2$ & $r_3$ & $r_4$ & $r_5$\\
  \hline
  $r_0$ & $0$  & $4$  & $6$  & $11$ & $11$ & $11$\\
  $r_1$ & $4$  & $0$  & $2$  & $11$ & $13$ & $11$\\
  $r_2$ & $6$  & $2$  & $0$  & $13$ & $11$ & $11$\\
  $r_3$ & $11$ & $11$ & $13$ & $0$  & $2$  & $12$\\
  $r_4$ & $11$ & $13$ & $11$ & $2$  & $0$  & $12$\\
  $r_5$ & $11$ & $11$ & $11$ & $12$ & $12$ & $0$\\
\end{tabular}
\caption{Distance matrix of the Ramon's repair distance}
\label{tabl:dist-matrix-short}

\end{table}

\label{ex:matrix-dist-ramon}
\end{example}

Please note that although we obtain distances between each pair of repairs, it does not mean that we can always place them in a 2D representation.

\begin{example}
Let us consider the distance matrix of Table \ref{tabl:dist-matrix-not-embed}, we can see that the points $A$, $B$ and $C$ form a right triangle with legs of length $1$ and the hypotenuse of length $\sqrt{2}$. The last point $D$ is at an equal distance of $1$ from each of these three points, but $D$ cannot be on the same plane as $A$, $B$ and $C$.

\begin{table}
\centering
\begin{tabular}{c|cccc}
      & $A$        & $B$ & $C$        & $D$ \\
  \hline
  $A$ & $0$        & $1$ & $\sqrt{2}$ & $1$ \\
  $B$ & $1$        & $0$ & $1$        & $1$ \\
  $C$ & $\sqrt{2}$ & $1$ & $0$        & $1$ \\
  $D$ & $1$        & $1$ & $1$        & $0$ \\
\end{tabular}
\caption{Example of a non 2D representation}
\label{tabl:dist-matrix-not-embed}
\end{table}

\label{ex:matrix-dist-not-embed}
\end{example}

In order to always produce a 2D representation, we thus used an approximation method called multi-dimensional scaling \cite{borg2005modern}. This method is able to find a set of points that minimise the difference between their distances and the target distance matrix by minimising the following stress function:

$$ Stress(x_1,\dots,x_N) = \sqrt{\sum_{i,j \in {1,\dots,N}} (d_{ij} - ||x_i - x_j||)^2} $$

\subsection{$\Dif$ Clustering function}
\label{sec:dif-clustering}

A usual way of making sense of large amount of data is to do clustering. By defining an appropriate notion of similarity on the data we are considering, we can get a high-level representation of the data. 
A clustering function is a function that group together similar repairs. 
In this paper, we used a clustering method called spectral clustering \cite{ng2002spectral} because it can take as input a similarity matrix that can easily be obtained from the distance matrix.

\begin{example}[Cont'd Example \ref{ex:matrix-dist-ramon}]
By using the spectral clustering on $\Emcon(\Kb)$, we obtain the following partition : $\{\{r_0, r_1, r_2\}, \{r_3, r_4\}, \{r_5\}\}$.
\label{ex:cluster-spectral}
\end{example}
\subsection{$\Dif$ Inference}
\label{sec:dif-inference}

\emph{Inconsistency-Tolerant Query Answering} is a challenging problem that received a lot of attention. We recall that we place ourselves in the context of OBDA, where the ontology is assumed to be fully reliable. 
In the following, we recall some of the most well-known inconsistency tolerant inferences. Let $\KB$ be a KB and $q$ be a boolean conjunctive query. $q$ is said to be \textbf{AR entailed} by $\Kb$ denoted by $\Kb \models_{AR} q$ if and only if for every $R \in \Emcon(\Kb), \Clo_{\R}(R) \models q$. $q$ is said to be \textbf{IAR entailed} by $\Kb$ denoted by $\Kb \models_{IAR} q$ if and only if $\Clo_{\R}\left(\bigcap\Emcon(\Kb)\right)  \models q$. $q$ is said to be \textbf{ICR entailed} by $\Kb$ denoted by $\Kb \models_{ICR} q$ if and only if $\bigcap\limits_{R \in \Emcon(\Kb)} \Clo_{\R}(R)  \models q$.

\begin{example}[Cont'd Example \ref{ex:list-repair}]
The query $q = \exists X\; baby(X)$ is AR entailed.
\end{example}

We propose to reuse AR, IAR, ICR by restricting them to a subset of repairs instead of the whole set of repairs. 

\begin{definition}[Restricted inference]
Let $x \in \{AR, IAR, ICR \}$. $\models^{R}_{x}$ denotes the restriction of $\models_{x}$ to the set of repairs $R$ instead of the whole set of repairs.
\end{definition}

\begin{definition}[$\Dif$ result]
The result of a DIF $\Dif = (d, \Clust, \models)$ on $\Kb$ for a query $q$ is a vector $(X_1, \dots, X_n)$ where $\{P_1, \dots, P_n\} = \Clust(\Emcon(\Kb), d)$ and for every $i \in \{1, \dots, n \}$, $X_i = True$ if $\Kb \models^{P_i} q$ and $False$ otherwise.
\end{definition}

\begin{example}[Cont'd Example \ref{ex:cluster-spectral}]
We have three clusters: $P_1 = \{ r_0, r_1, r_2 \}, P_2 = \{ r_3, r_4 \}$ and $P_3 = \{ r_5 \}$. The result of $\Dif = ( d, \Clust, \models)$ on the query $q = \exists X (baby(X) \wedge get\_ill(X)) $ is $(True, True, False)$
\end{example}

\section{Application \& Discussion}

The usual inconsistency-tolerant inference approaches are sometimes unintelligible and not straightforward for the end-user as they implement complex repairing strategies. We propose an automatic tool for visualising and performing a personalised query answering.

\begin{figure}[!h]
\centering
\includegraphics[width=10cm]{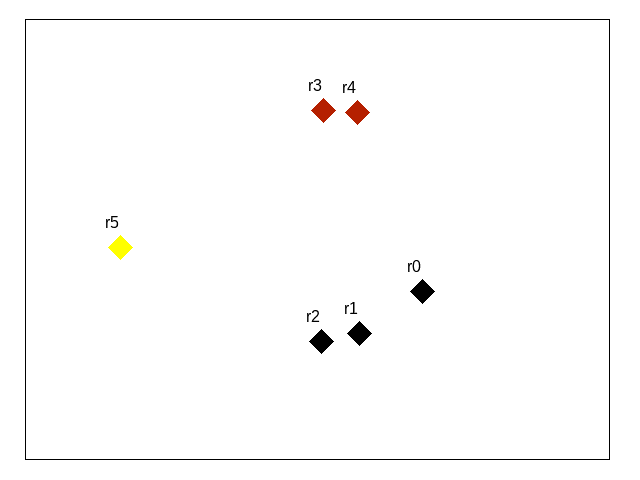}
\caption{Plot of the similarity between repairs}
\label{fig-representation-points}
\end{figure}

As discussed in Section \ref{sec:dif-distance}, we cannot always find an exact 2D representation that fits the distance matrix. In order to obtain the representation of Figure \ref{fig-representation-points}, we use the multidimensional scaling method \cite{borg_modern_2003} to obtain an embedding of the repairs in a 2D space that tries to best respect the distance matrix. Then, using the spectral clustering method, we obtain a $K$ colouring of the repairs corresponding to the several clusters.
In our workflow, the end-user is able to:
\begin{enumerate}
\item have a intuitive visualisation of the repairs and the syntactic distance between them,
\item get the answer to a query on a particular set of repair by clicking on the corresponding cluster and,
\item have a more personalised answer by manually selecting the repairs one wants to query on.
\end{enumerate}

The code for computing the repairs and the distance framework can be found in the following repositories: \texttt{https://gite.lirmm.fr/yun/Dagger} and \texttt{https://gite.lirmm.fr/proute/repairs-visualisation}

\bibliographystyle{unsrt}  

 \bibliography{bruno}
 
\end{document}